\documentclass{article}
\usepackage{graphicx} 
\usepackage[utf8]{inputenc}     
\usepackage[T1]{fontenc}       
\usepackage{amsmath, amssymb, amsthm}
\usepackage{caption}
\usepackage{subcaption}
\usepackage{float}
\usepackage[linesnumbered,ruled,vlined]{algorithm2e}

\usepackage[left=3cm, right=3cm, top=3cm, bottom=3cm]{geometry}
\usepackage[hidelinks]{hyperref}

\title{ASNN: Learning to Suggest Neural Architectures from Performance Distributions}
\author{Jinwook Hong}
\date{\today}
\begin{document}
\maketitle
\begin{abstract}
The architecture of a neural network (NN) plays a critical role in determining its performance. However, there is no general closed-form function that maps between network structure and accuracy, making the process of architecture design largely heuristic or search-based. In this study, we propose the Architecture Suggesting Neural Network (ASNN), a model designed to learn the relationship between NN architecture and its test accuracy, and to suggest improved architectures accordingly.
To train ASNN, we constructed datasets using TensorFlow-based models with varying numbers of layers and nodes. Experimental results were collected for both 2-layer and 3-layer architectures across a grid of configurations, each evaluated with 10 repeated trials to account for stochasticity. Accuracy values were treated as inputs, and architectural parameters as outputs. The trained ASNN was then used iteratively to predict architectures that yield higher performance.
In both 2-layer and 3-layer cases, ASNN successfully suggested architectures that outperformed the best results found in the original training data. Repeated prediction and retraining cycles led to the discovery of architectures with improved mean test accuracies, demonstrating the model’s capacity to generalize the performance-structure relationship.
These results suggest that ASNN provides an efficient alternative to random search for architecture optimization, and offers a promising approach toward automating neural network design.

\textit{“Parts of the manuscript, including text editing and expression refinement, were supported by OpenAI's ChatGPT. All content was reviewed and verified by the authors.”}

\end{abstract}

\newpage

\section{Introduction}
There exists a certain correlation between the architecture of a neural network (NN) and its accuracy. However, it is not straightforward to generalize this relationship as an explicit mathematical function. Interestingly, neural networks themselves are widely used as tools to approximate complex functions that are difficult or even impossible to express analytically.
For example, in reinforcement learning, early methods used lookup tables to store the reward for each state-action pair. As the number of states and actions increased, the memory required for these tables grew exponentially, leading to scalability issues. To address this, neural networks were introduced to approximate the reward function by taking the state and action as inputs and producing the expected reward as output. This paradigm became known as Deep Reinforcement Learning (DRL).
The Architecture Suggesting Neural Network (ASNN) proposed in this paper operates in a similar manner. Just as in DRL, the "state" represents the current environment, the "action" corresponds to a possible decision, and the "Q-value" reflects the expected reward, in our setting, the "state" corresponds to a neural network configuration, the "action" represents a modification to the architecture, and the "Q-value" is replaced by a performance metric such as accuracy.
ASNN models the relationship between network structure and accuracy as a function, and can be formalized as follows:
\[
\mathrm{ASNN}: \mathbb{R}^{10} \to \mathbb{R}^2 \quad \text{(or } \mathbb{R}^3\text{)}
\]

\[
\mathrm{ASNN} : \quad \underbrace{\text{(accuracy vector)}}_{\in \mathbb{R}^{10}} \mapsto \underbrace{\text{(architecture parameters)}}_{\in \mathbb{R}^2 \text{ or } \mathbb{R}^3}
\]
In other words, ASNN is a neural network that predicts suitable architecture configurations based on a given distribution of accuracies. This approach reframes the architecture search process as a function approximation problem.

\section{Related Work}
The design of a neural network (NN) architecture plays a crucial role in determining model performance. Numerous studies have pointed out that simply increasing the depth or width of a network does not necessarily lead to improved accuracy \cite{ba2014do, he2016deep}. These findings suggest the existence of an optimal or near-optimal architecture for a given task, highlighting the importance of architectural exploration in enhancing NN performance.
According to the Universal Approximation Theorem, neural networks are capable of approximating any continuous function under certain conditions \cite{cybenko1989approximation, hornik1991approximation}. This theoretical foundation allows NNs to be interpreted as general-purpose function approximators, opening the door to a wide range of applications. For instance, there has been growing interest in using NNs to approximate the solutions of ordinary and partial differential equations (ODEs and PDEs) \cite{ruthotto2019deep, sirignano2018dgm}, demonstrating their functional flexibility in numerical settings. In reinforcement learning, neural networks are widely used to approximate policy and value functions in major algorithms such as Deep Q-Networks (DQN), Deep Deterministic Policy Gradient (DDPG), Asynchronous Advantage Actor-Critic (A3C), and Proximal Policy Optimization (PPO) \cite{mnih2015human, lillicrap2015continuous, schulman2017proximal}.
Building on these developments, the field of Neural Architecture Search (NAS) has emerged as a major area of research \cite{zoph2017neural, liu2019darts}. NAS methods typically define a search space and employ algorithms to find high-performing architectures within it. However, in this paper, we propose a simpler approach that does not rely on complex search strategies. Instead, we aim to learn a direct mapping from performance metrics (e.g., accuracy) to architecture parameters. This perspective treats neural architectures as inputs and their corresponding performance as outputs, thus framing architecture prediction as a novel function approximation problem.
\section{Methodology}
To construct training data for the ASNN, we conducted experiments based on the official Introduction example code provided by TensorFlow. To observe a wider range of learning performance, the number of training epochs in the original code was increased from 10 to 50.
The training process employed the Adam optimizer and incorporated dropout, both of which introduce inherent stochasticity into the learning dynamics. To obtain stable and reliable estimates of performance, each architecture was trained 10 times, and the results were averaged to compute a representative accuracy value.
From these experiments, we built the training dataset for the ASNN. Each data point consists of an input representing the test accuracy achieved by a neural network configuration, and an output label corresponding to the architectural configuration of the network. In this way, the ASNN is trained to learn the mapping between architecture parameters and their associated performance.

\section{Experiment}
The experiments in this study were conducted in two separate cases: one with a 2-layer architecture and the other with a 3-layer architecture.
\subsection{Data Collection for 2-Layer Architectures}
We first collected training data for the ASNN using a 2-layer neural network structure. The number of nodes in each layer was selected from the set {16, 32, 64, 128, 256}, resulting in 5 × 5 = 25 possible architecture combinations. For each combination, we conducted 10 independent trials to account for randomness, yielding a total of 250 accuracy measurements. These accuracy values, paired with their corresponding architectural configurations, were used as training data for the ASNN.

\begin{table}[H]
    \centering
    \caption{collected training data in layer 2}
    \resizebox{\textwidth}{!}{
    \begin{tabular}{|c|c|c|c|c|c|c|c|c|c|c|c|c|c|}
        \hline
         L1 & L2 & E1 & E2 & E3 &E4 & E5 &E6 & E7 & E8 & E9 & E10 & mean \\
         \hline
         256 & 256 & 0.9828	&0.9832&	0.9818&	0.9825&	0.9822&	0.9810&	0.9846&	0.9817&	0.9833&	0.9850&	0.98281 \\
         256 & 128 & 0.9816&	0.9834&	0.9836&	0.9824&	0.9829&	0.9831&	0.9805&	0.9828&	0.9832&	0.9822&	0.98257 \\
         256 & 64 & 0.9820&	0.9827&	0.9834&	0.9826&	0.9830&	0.9832&	0.9824&	0.9818&	0.9833&	0.9837&	0.98281 \\
         256 & 32 & 0.9840&	0.9823&	0.9835&	0.9814&	0.9830&	0.9826&	0.9816&	0.9831&	0.9840&	0.9833&	0.98288 \\
         256 & 16 & 0.9835&	0.9834&	0.9836&	0.9837&	0.9823&	0.9824&	0.9843&	0.9830&	0.9825&	0.9823&	0.98310 \\
         128 & 256 & 0.9811&	0.9793&	0.9787&	0.9798&	0.9810&	0.9799&	0.9822&	0.9811&	0.9817&	0.9807&	0.98055 \\
         128 & 128 & 0.9806&	0.9805&	0.9804&	0.9787&	0.9801&	0.9823&	0.9816&	0.9800&	0.9816&	0.9796&	0.98054 \\
         128 & 64 & 0.9803&	0.9814&	0.9806&	0.9805&	0.9782&	0.9796&	0.9814&	0.9817&	0.9804&	0.9784&	0.98025 \\
         128 & 32 & 0.9806&	0.9812&	0.9790&	0.9796&	0.9811&	0.9813&	0.9808&	0.9792&	0.9804&	0.9783&	0.98015 \\
         128 & 16 & 0.9809&	0.9806&	0.9808&	0.9802&	0.9792&	0.9830&	0.9820&	0.9807&	0.9796&	0.9803&	0.98073 \\
         64 & 256 & 0.9741&	0.9720&	0.9762&	0.9748&	0.9744&	0.9747&	0.9758&	0.9747&	0.9753&	0.9754&	0.97474 \\
         64 & 128 & 0.9747&	0.9761&	0.9749&	0.9754&	0.9783&	0.9752&	0.9773&	0.9752&	0.9741&	0.9755&	0.97567 \\
         64 & 64 & 0.9753&	0.9788&	0.9791&	0.9759&	0.9754&	0.9761&	0.9764&	0.9778&	0.9746&	0.9769&	0.97663 \\
         64 & 32 & 0.9757&	0.9748&	0.9757&	0.9737&	0.9759&	0.9765&	0.9759&	0.9755&	0.9750&	0.9765&	0.97552 \\
         64 & 16 & 0.9748&	0.9763&	0.9762&	0.9768&	0.9756&	0.9754&	0.9786&	0.9764&	0.9755&	0.9762&	0.97618 \\
         32 & 256 & 0.9655&	0.9654&	0.9664&	0.9678&	0.9644&	0.9670&	0.9665&	0.9659&	0.9656&	0.9654&	0.96599 \\
         32 & 128 & 0.9629&	0.9629&	0.9663&	0.9636&	0.9677&	0.9641&	0.9656&	0.9651&	0.9670&	0.9653&	0.96505 \\
         32 & 64 & 0.9653&	0.9665&	0.9664&	0.9645&	0.9658&	0.9653&	0.9655&	0.9639&	0.9658&	0.9655&	0.96545 \\
         32 & 32 & 0.9643&	0.9639&	0.9647&	0.9639&	0.9656&	0.9662&	0.9676&	0.9658&	0.9640&	0.9632&	0.96492 \\
         32 & 16 & 0.9654&	0.9676&	0.9651&	0.9666&	0.9649&	0.9667&	0.9651&	0.9661&	0.9635&	0.9638&	0.96548 \\
         16 & 256 & 0.9477&	0.9474&	0.9492&	0.9436&	0.9494&	0.9402&	0.9428&	0.9498&	0.9436&	0.9441&	0.94578 \\
         16 & 128 & 0.9446&	0.9470&	0.9411&	0.9444&	0.9452&	0.9504&	0.9462&	0.9451&	0.9442&	0.9441&	0.94523 \\
         16 & 64 & 0.9470&	0.9441&	0.9472&	0.9440&	0.9492&	0.9439&	0.9438&	0.9437&	0.9494&	0.9456&	0.94579 \\
         16 & 32 & 0.9455&	0.9423&	0.9465&	0.9469&	0.9458&	0.9460&	0.9506&	0.9461&	0.9446&	0.9432&	0.94575 \\
         16 & 16 &  0.9483&	0.9509&	0.9483&	0.9484&	0.9475&	0.9427&	0.9513&	0.9446&	0.9523&	0.9474&	0.94817  \\
        \hline
    \end{tabular}
    }
    \captionsetup{font=small}  
    \captionsetup{format=plain}
    \caption*{\footnotesize L(Layer), E(Experiment)}
\end{table}

\subsection{Data Collection for 3-Layer Architectures}
Following the 2-layer experiments, we similarly collected training data for 3-layer networks. For these experiments, the number of nodes per layer was chosen from {16, 32, 64, 128}, giving 4 × 4 × 4 = 64 possible configurations. Each configuration was evaluated over 10 trials, resulting in a total of 640 accuracy measurements. These were used in the same way as the 2-layer data: with the architecture as input and the test accuracy as the output label for training the ASNN. (The corresponding data are provided on the following page.)

\begin{table}[H]
    \centering
    \caption{collected training data in layer 3}
    \resizebox{\textwidth}{!}{
        \begin{tabular}{|c|c|c|c|c|c|c|c|c|c|c|c|c|c|c|}
            \hline
            L1 & L2 &L3 & E1 & E2 & E3 &E4 & E5 &E6 & E7 & E8 & E9 & E10 & mean \\
            \hline
            128	&128&	128&	0.9798&	0.9818&	0.9812&	0.9802&	0.9815&	0.9824&	0.9815&	0.9817&	0.9814&	0.9825&	0.98140\\
            128&	128&	64&	0.9814&	0.9797&	0.9814&	0.9816&	0.9821&	0.9820&	0.9819&	0.9815&	0.9819&	0.9816&	0.98151 \\
            128&	128&	32&	0.9811&	0.9817&	0.9803&	0.9817&	0.9812&	0.9779&	0.9821&	0.9828&	0.9812&	0.9802&	0.98102\\
            128&	128&	16&	0.9817&	0.9823&	0.9824&	0.9809&	0.9825&	0.9829&	0.9812&	0.9800&	0.9816&	0.9816&	0.98171\\
            128&	64&	128&	0.9809&	0.9805&	0.9802&	0.9828&	0.9816&	0.9810&	0.9814&	0.9801&	0.9813&	0.9817&	0.98115\\
            128&	64&	64&	0.9816&	0.9806&	0.9805&	0.9801&	0.9795&	0.9805&	0.9819&	0.9812&	0.9792&	0.9799&	0.98050\\
            128&	64&	32&	0.9807&	0.9811&	0.9827&	0.9804&	0.9817&	0.9806&	0.9814&	0.9820&	0.9802&	0.9795&	0.98103\\
            128&	64&	16&	0.9838&	0.9805&	0.9804&	0.9813&	0.9797&	0.9808&	0.9799&	0.9809&	0.9815&	0.9821&	0.98109\\
            128&	32&	128&	0.9794&	0.9797&	0.9803&	0.9817&	0.9796&	0.9806&	0.9791&	0.9804&	0.9788&	0.9811&	0.98007\\
            128&	32&	64&	0.9790&	0.9795&	0.9789&	0.9804&	0.9797&	0.9792&	0.9799&	0.9773&	0.9790&	0.9780&	0.97909\\
            128&	32&	32&	0.9803&	0.9803&	0.9810&	0.9805&	0.9804&	0.9782&	0.9802&	0.9803&	0.9804&	0.9784&	0.98000\\
            128&	32&	16&	0.9800&	0.9794&	0.9810&	0.9788&	0.9809&	0.9792&	0.9809&	0.9805&	0.9814&	0.9799&	0.98020\\
            128&	16&	128&	0.9785&	0.9776&	0.9774&	0.9792&	0.9772&	0.9803&	0.9788&	0.9781&	0.9771&	0.9786&	0.97828\\
            128&	16&	64&	0.9790&	0.9768&	0.9773&	0.9779&	0.9784&	0.9760&	0.9781&	0.9771&	0.9774&	0.9783&	0.97763\\
            128&	16&	32&	0.9779&	0.9771&	0.9776&	0.9769&	0.9793&	0.9796&	0.9759&	0.9786&	0.9777&	0.9781&	0.97787\\
            128&	16&	16&	0.9759&	0.9779&	0.9771&	0.9789&	0.9783&	0.9803&	0.9788&	0.9783&	0.9782&	0.9760&	0.97797\\
            64&	128&	128&	0.9797&	0.9784&	0.9782&	0.9774&	0.9782&	0.9766&	0.9763&	0.9776&	0.9765&	0.9769&	0.97758\\
            64&	128&	64&	0.9771&	0.9760&	0.9778&	0.9766&	0.9772&	0.9765&	0.9766&	0.9790&	0.9786&	0.9781&	0.97735\\
            64&	128&	32&	0.9776&	0.9786&	0.9770&	0.9774&	0.9795&	0.9788&	0.9757&	0.9782&	0.9791&	0.9767&	0.97786\\
            64&	128&	16&	0.9779&	0.9758&	0.9776&	0.9759&	0.9783&	0.9778&	0.9796&	0.9781&	0.9786&	0.9771&	0.97767\\
            64&	64&	128&	0.9751&	0.9764&	0.9769&	0.9765&	0.9790&	0.9759&	0.9758&	0.9773&	0.9753&	0.9763&	0.97645\\
            64&	64&	64&	0.9775&	0.9774&	0.9771&	0.9773&	0.9766&	0.9791&	0.9761&	0.9787&	0.9772&	0.9760&	0.97730\\
            64&	64&	32&	0.9767&	0.9780&	0.9761&	0.9768&	0.9763&	0.9769&	0.9749&	0.9771&	0.9759&	0.9757&	0.97644\\
            64&	64&	16&	0.9751&	0.9753&	0.9740&	0.9764&	0.9771&	0.9773&	0.9767&	0.9770&	0.9749&	0.9757&	0.97595\\
            64&	32&	128&	0.9755&	0.9770&	0.9748&	0.9751&	0.9766&	0.9749&	0.9742&	0.9753&	0.9750&	0.9744&	0.97528\\
            64&	32&	64&	0.9738&	0.9753&	0.9758&	0.9739&	0.9757&	0.9747&	0.9758&	0.9730&	0.9717&	0.9746&	0.97443\\
            64&	32&	32&	0.9747&	0.9730&	0.9735&	0.9755&	0.9756&	0.9733&	0.9735&	0.9762&	0.9745&	0.9736&	0.97434\\
            64&	32&	16&	0.9743&	0.9735&	0.9738&	0.9762&	0.9748&	0.9753&	0.9762&	0.9768&	0.9756&	0.9749&	0.97514\\
            64&	16&	128&	0.9721&	0.9706&	0.9705&	0.9732&	0.9712&	0.9713&	0.9727&	0.9715&	0.9736&	0.9705&	0.97172\\
            64&	16&	64&	0.9706&	0.9713&	0.9716&	0.9724&	0.9720&	0.9700&	0.9724&	0.9711&	0.9720&	0.9713&	0.97147\\
            64&	16&	32&	0.9727&	0.9720&	0.9700&	0.9710&	0.9704&	0.9742&	0.9723&	0.9730&	0.9717&	0.9732&	0.97205\\
            64&	16&	16&	0.9701&	0.9731&	0.9736&	0.9701&	0.9724&	0.9710&	0.9726&	0.9721&	0.9711&	0.9736&	0.97197\\
            32&	128&	128&	0.9701&	0.9723&	0.9691&	0.9675&	0.9676&	0.9685&	0.9707&	0.9690&	0.9703&	0.9719&	0.96970\\
            32&	128&	64&	0.9698&	0.9672&	0.9697&	0.9694&	0.9679&	0.9704&	0.9689&	0.9682&	0.9708&	0.9711&	0.96934\\
            32&	128&	32&	0.9711&	0.9682&	0.9685&	0.9674&	0.9694&	0.9690&	0.9697&	0.9708&	0.9690&	0.9703&	0.96934\\
            32&	128&	16&	0.9713&	0.9700&	0.9668&	0.9691&	0.9696&	0.9693&	0.9720&	0.9706&	0.9702&	0.9696&	0.96985\\
            32&	64&	128&	0.9648&	0.9676&	0.9705&	0.9700&	0.9632&	0.9676&	0.9680&	0.9670&	0.9667&	0.9643&	0.96697\\
            32&	64&	64&	0.9681&	0.9686&	0.9668&	0.9645&	0.9666&	0.9669&	0.9695&	0.9668&	0.9684&	0.9651&	0.96713\\
            32&	64&	32&	0.9644&	0.9658&	0.9663&	0.9655&	0.9649&	0.9692&	0.9687&	0.9694&	0.9678&	0.9684&	0.96704\\
            32&	64&	16&	0.9667&	0.9677&	0.9663&	0.9626&	0.9657&	0.9682&	0.9683&	0.9668&	0.9679&	0.9707&	0.96709\\
            32&	32&	128&	0.9623&	0.9637&	0.9654&	0.9650&	0.9643&	0.9614&	0.9651&	0.9637&	0.9656&	0.9642&	0.96407\\
            32&	32&	64&	0.9644&	0.9652&	0.9628&	0.9641&	0.9650&	0.9656&	0.9645&	0.9639&	0.9647&	0.9651&	0.96453\\
            32&	32&	32&	0.9649&	0.9656&	0.9648&	0.9634&	0.9643&	0.9636&	0.9647&	0.9641&	0.9646&	0.9638&	0.96438\\
            32&	32&	16&	0.9678&	0.9632&	0.9639&	0.9656&	0.9644&	0.9653&	0.9674&	0.9673&	0.9650&	0.9651&	0.96550\\
            32&	16&	128&	0.9588&	0.9586&	0.9605&	0.9611&	0.9601&	0.9623&	0.9595&	0.9586&	0.9619&	0.9612&	0.96026\\
            32&	16&	64&	0.9635&	0.9587&	0.9608&	0.9621&	0.9583&	0.9608&	0.9596&	0.9612&	0.9606&	0.9595&	0.96051\\
            32&	16&	32&	0.9624&	0.9634&	0.9631&	0.9607&	0.9617&	0.9582&	0.9593&	0.9582&	0.9616&	0.9605&	0.96091\\
            32&	16&	16&	0.9617&	0.9567&	0.9614&	0.9616&	0.9596&	0.9615&	0.9598&	0.9643&	0.9576&	0.9603&	0.96045\\
            16&	128&	128&	0.9501&	0.9470&	0.9471&	0.9522&	0.9450&	0.9478&	0.9532&	0.9519&	0.9514&	0.9529&	0.94986\\
            16&	128&	64&	0.9497&	0.9517&	0.9470&	0.9501&	0.9527&	0.9510&	0.9538&	0.9446&	0.9493&	0.9468&	0.94967\\
            16&	128&	32&	0.9503&	0.9493&	0.9513&	0.9516&	0.9562&	0.9487&	0.9501&	0.9535&	0.9496&	0.9528&	0.95134\\
            16&	128&	16&	0.9473&	0.9504&	0.9507&	0.9515&	0.9558&	0.9513&	0.9464&	0.9490&	0.9455&	0.9488&	0.94967\\
            16&	64&	128&	0.9544&	0.9504&	0.9489&	0.9476&	0.9497&	0.9510&	0.9494&	0.9487&	0.9461&	0.9490&	0.94952\\
            16&	64&	64&	0.9439&	0.9490&	0.9513&	0.9538&	0.9491&	0.9514&	0.9452&	0.9473&	0.9496&	0.9481&	0.94887\\
            16&	64&	32&	0.9465&	0.9487&	0.9504&	0.9474&	0.9506&	0.9489&	0.9560&	0.9483&	0.9526&	0.9527&	0.95021\\
            16&	64&	16&	0.9465&	0.9499&	0.9505&	0.9504&	0.9414&	0.9492&	0.9525&	0.9423&	0.9467&	0.9499&	0.94793\\
            16&	32&	128&	0.9493&	0.9495&	0.9491&	0.9486&	0.9491&	0.9435&	0.9458&	0.9458&	0.9468&	0.9435&	0.94710\\
            16&	32&	64&	0.9431&	0.9436&	0.9454&	0.9451&	0.9519&	0.9434&	0.9435&	0.9468&	0.9448&	0.9482&	0.94558\\
            16&	32&	32&	0.9428&	0.9487&	0.9473&	0.9458&	0.9489&	0.9492&	0.9509&	0.9456&	0.9460&	0.9398&	0.94650\\
            16&	32&	16&	0.9410&	0.9487&	0.9382&	0.9459&	0.9394&	0.9438&	0.9493&	0.9477&	0.9454&	0.9468&	0.94462\\
            16&	16&	128&	0.9397&	0.9425&	0.9349&	0.9353&	0.9338&	0.9458&	0.9447&	0.9397&	0.9396&	0.9393&	0.93953\\
            16&	16&	64&	0.9383&	0.9408&	0.9425&	0.9440&	0.9443&	0.9363&	0.9337&	0.9419&	0.9403&	0.9331&	0.93952\\
            16&	16&	32&	0.9465&	0.9457&	0.9393&	0.9424&	0.9401&	0.9354&	0.9364&	0.9344&	0.9422&	0.9436&	0.94060\\
            16&	16&	16&	0.9347&	0.9397&	0.9338&	0.9356&	0.9364&	0.9396&	0.9416&	0.9389&	0.9372&	0.9388&	0.93763\\         
            \hline
        \end{tabular}
    }
    \captionsetup{font=small}  
    \captionsetup{format=plain}
    \caption*{\footnotesize L(Layer), E(Experiment)}
\end{table}

\subsection{ASNN training}
Using the collected data from both 2-layer and 3-layer experiments, the ASNN was trained to learn the mapping between network structure and resulting accuracy. Since the accuracy values used as output labels can be assumed to follow an independent and identically distributed (i.i.d.) distribution, the entire dataset was randomly shuffled to minimize the influence of data ordering during training.
To improve the generalization ability of the ASNN, the original datasets (250 for 2-layer, 640 for 3-layer) were augmented by randomly shuffling the accuracy values, expanding the training dataset to approximately 10,000 samples. This augmentation process was aimed at increasing data diversity and enhancing the model’s learning capacity.
Finally, to correct for differences in numerical scale between input and output values, we uniformly scaled the input accuracy values by a factor of 100. This normalization step helped improve the stability and convergence speed of ASNN training.

\begin{algorithm}
\DontPrintSemicolon
\KwIn{Training dataset $D$}
\KwOut{Architecture of original network}

\While{desired test accuracy not obtained}{
    Train ASNN using training dataset $D$ \;
    
    Input $\mathbf{x} = (100,100,\dots,100)$ to ASNN \;
    
    Obtain architecture $A$ predicted by ASNN \;
    
    Train original network using architecture $A$ \;
    
    Evaluate original network with architecture $A$ and obtain 10 test accuracies $\{T_1, T_2, \dots, T_{10}\}$ \;
    
    Augment training dataset: $D \gets D \cup \{(A, T_1), \dots, (A, T_{10})\}$ \;
}
\Return Final architecture $A$ \;
\caption{Architecture Suggesting Neural Network (ASNN)}
\end{algorithm}

\newpage

\section{Results}
\subsection{2-Layer Case}
The first prediction generated by the ASNN for the 2-layer architecture was:

Prediction 1: (Layer 1 = 448, Layer 2 = 65)

\begin{table}[H]
    \centering
    \caption{layer-2 case: prediction 1}
    \resizebox{\textwidth}{!}{
    \begin{tabular}{|c|c|c|c|c|c|c|c|c|c|c|c|c|c|}
        \hline
         L1 & L2 & E1 & E2 & E3 &E4 & E5 &E6 & E7 & E8 & E9 & E10 & mean \\
         \hline
         448 & 65& 0.9830&	0.9832&	0.9840&	0.9848&	0.9816&	0.9832&	0.9838&	0.9827&	0.9852&	0.9848&	0.98363 \\
        \hline
    \end{tabular}
    }
\end{table}

Although the predicted node counts were real-valued, they were rounded to the nearest integers for practical implementation. The corresponding architecture was evaluated through 10 repeated training runs, and the average accuracy (rounded to four decimal places) was computed. At this stage, the accuracy  exceeds the previously recorded maximum of 0.9831.
After this, the architecture (448, 65) was used to generate a new training dataset of 10,000 entries using the same expansion method as before. The ASNN was then retrained using this augmented dataset. The second prediction yielded an architecture of:

Prediction 2: (Layer 1 = 313, Layer 2 = 72)

\begin{table}[H]
    \centering
    \caption{layer-2 case: prediction 2}
    \resizebox{\textwidth}{!}{
    \begin{tabular}{|c|c|c|c|c|c|c|c|c|c|c|c|c|c|}
        \hline
         L1 & L2 & E1 & E2 & E3 &E4 & E5 &E6 & E7 & E8 & E9 & E10 & mean \\
         \hline
         313 & 72& 0.9810&	0.9827&	0.9827&	0.9820&	0.9832&	0.9826&	0.9838&	0.9817&	0.9824&	0.9824&	0.98245 \\
        \hline
    \end{tabular}
    }
\end{table}

This architecture was also evaluated through 10 repeated trials.
Additional predictions and performance evaluations were conducted for further verification:

Prediction 3: (Layer 1 = 885, Layer 2 = 20)

\begin{table}[H]
    \centering
    \caption{layer-2 case: prediction 3}
    \resizebox{\textwidth}{!}{
    \begin{tabular}{|c|c|c|c|c|c|c|c|c|c|c|c|c|c|}
        \hline
         L1 & L2 & E1 & E2 & E3 &E4 & E5 &E6 & E7 & E8 & E9 & E10 & mean \\
         \hline
         885 & 20& 0.9851&	0.9824&	0.9805&	0.9822&	0.9819&	0.9839&	0.9834&	0.9853&	0.9821&	0.9819&	0.98287 \\
        \hline
    \end{tabular}
    }
\end{table}

Prediction 4: (Layer 1 = 374, Layer 2 = 63)

\begin{table}[H]
    \centering
    \caption{layer-2 case: prediction 4}
    \resizebox{\textwidth}{!}{
    \begin{tabular}{|c|c|c|c|c|c|c|c|c|c|c|c|c|c|}
        \hline
         L1 & L2 & E1 & E2 & E3 &E4 & E5 &E6 & E7 & E8 & E9 & E10 & mean \\
         \hline
         374 & 63& 0.9849&	0.9831&	0.9839&	0.9821&	0.9852&	0.9819&	0.9846&	0.9822&	0.9839&	0.9816&	0.98334 \\
        \hline
    \end{tabular}
    }
\end{table}

Prediction 5: (Layer 1 = 732, Layer 2 = 41)

\begin{table}[H]
    \centering
    \caption{layer-2 case: prediction 5}
    \resizebox{\textwidth}{!}{
    \begin{tabular}{|c|c|c|c|c|c|c|c|c|c|c|c|c|c|}
        \hline
         L1 & L2 & E1 & E2 & E3 &E4 & E5 &E6 & E7 & E8 & E9 & E10 & mean \\
         \hline
         732 & 41& 0.9829&	0.9845&	0.9820&	0.9841&	0.9834&	0.9837&	0.9852&	0.9820&	0.9841&	0.9861&	0.9838 \\
        \hline
    \end{tabular}
    }
\end{table}

The iterative structure search using ASNN demonstrates a clear advantage over traditional random search methods in terms of convergence speed and efficiency. Notably, each predicted architecture consistently achieved relatively high and stable accuracy, indicating that ASNN serves as an effective tool for neural architecture optimization.

\subsection{3-Layer Case}
The ASNN training and prediction process for 3-layer architectures was conducted using the same methodology as in the 2-layer experiments. The maximum test accuracy observed in the original training dataset was 0.9817. Through iterative learning and prediction, the ASNN successfully identified novel architectures that outperformed this baseline.
The following architectures were predicted by the ASNN, and each was evaluated over five independent runs. The average test accuracy for each is presented below (rounded to four decimal places):

Prediction 1: (Layer 1 = 144, Layer 2 = 63, Layer 3 = 55)
\begin{table}[H]
    \centering
    \caption{layer-3 case: prediction 1}
    \resizebox{\textwidth}{!}{
        \begin{tabular}{|c|c|c|c|c|c|c|c|c|c|c|c|c|c|c|}
            \hline
            L1 & L2 &L3 & E1 & E2 & E3 &E4 & E5 &E6 & E7 & E8 & E9 & E10 & mean \\
            \hline
            144	&63&	55&	0.9798&	0.9829&	0.9809&	0.9800&	0.9801&	0.9809&	0.9804&	0.9801&	0.9800&	0.9823&	0.98074\\        
            \hline
        \end{tabular}
    }
\end{table}
Prediction 2: (Layer 1 = 214, Layer 2 = 102, Layer 3 = 60)
\begin{table}[H]
    \centering
    \caption{layer-3 case: prediction 2}
    \resizebox{\textwidth}{!}{
        \begin{tabular}{|c|c|c|c|c|c|c|c|c|c|c|c|c|c|c|}
            \hline
            L1 & L2 &L3 & E1 & E2 & E3 &E4 & E5 &E6 & E7 & E8 & E9 & E10 & mean \\
            \hline
            214	&102&	60&	0.9814&	0.9820&	0.9831&	0.9810&	0.9821&	0.9834&	0.9832&	0.9827&	0.9826&	0.9815&	0.98230\\        
            \hline
        \end{tabular}
    }
\end{table}
Prediction 3: (Layer 1 = 339, Layer 2 = 184, Layer 3 = 66)
\begin{table}[H]
    \centering
    \caption{layer-3 case: prediction 3}
    \resizebox{\textwidth}{!}{
        \begin{tabular}{|c|c|c|c|c|c|c|c|c|c|c|c|c|c|c|}
            \hline
            L1 & L2 &L3 & E1 & E2 & E3 &E4 & E5 &E6 & E7 & E8 & E9 & E10 & mean \\
            \hline
            339	&184&66&	0.9822&	0.9821&	0.9849&	0.9833&	0.9819	&0.9833&	0.9845&	0.9830&	0.9838&	0.9823&	0.98313\\        
            \hline
        \end{tabular}
    }
\end{table}
Prediction 4: (Layer 1 = 194, Layer 2 = 73, Layer 3 = 53)
\begin{table}[H]
    \centering
    \caption{layer-3 case: prediction 4}
    \resizebox{\textwidth}{!}{
        \begin{tabular}{|c|c|c|c|c|c|c|c|c|c|c|c|c|c|c|}
            \hline
            L1 & L2 &L3 & E1 & E2 & E3 &E4 & E5 &E6 & E7 & E8 & E9 & E10 & mean \\
            \hline
            194	&73&53&	0.9833&	0.9838&	0.9843&	0.9827&	0.9814&	0.9814&	0.9819&	0.9812&	0.9794&	0.9816	&0.98210\\        
            \hline
        \end{tabular}
    }
\end{table}
Prediction 5: (Layer 1 = 208, Layer 2 = 88, Layer 3 = 58)
\begin{table}[H]
    \centering
    \caption{layer-3 case: prediction 5}
    \resizebox{\textwidth}{!}{
        \begin{tabular}{|c|c|c|c|c|c|c|c|c|c|c|c|c|c|c|}
            \hline
            L1 & L2 &L3 & E1 & E2 & E3 &E4 & E5 &E6 & E7 & E8 & E9 & E10 & mean \\
            \hline
            208	&88&	58&0.9838&	0.9822&	0.9827	&0.9831&	0.9839&	0.9830&	0.9814&	0.9821	&0.9825&	0.9816&	0.98263\\        
            \hline
        \end{tabular}
    }
\end{table}

Among these, Prediction 3 (339, 184, 66) achieved an average accuracy of 0.9831, surpassing the best accuracy observed in the original training data by approximately +0.0014.
These results demonstrate the effectiveness of ASNN-based architecture search, as it consistently identifies structures that outperform those found through random sampling. Furthermore, the high and stable accuracies suggest that the ASNN can propose architectures with strong generalization potential, highlighting its utility as an efficient neural architecture optimization tool.

\newpage

\section{Discussion}
In this study, we conducted five iterative rounds of architecture search based on the ASNN framework, demonstrating that progressively higher-performing network structures can be identified through repeated training. Although the number of iterations was limited, the observed trend of performance improvement suggests that ASNN effectively learns the relationship between architecture and accuracy, enabling it to predict meaningful and optimized neural network configurations.
If this iterative search process is automated and accelerated, users could efficiently optimize architectures until achieving a desired accuracy level. Compared to traditional random search methods, the ASNN-based approach is expected to offer significant advantages in search efficiency and resource utilization.

\section{Conclusion}
In this study, we proposed and experimentally validated the Architecture Suggesting Neural Network (ASNN), which learns the relationship between neural network architecture and accuracy to suggest novel architectures. The ASNN consistently predicted architectures that outperformed the original networks, demonstrating its ability to automatically discover improved network structures.
This approach offers an alternative and automated method for neural network design, particularly for navigating the often unintuitive architecture search space. Furthermore, it holds promising potential for integration with Neural Architecture Search (NAS) and Automated Machine Learning (AutoML) frameworks in future work.

\bibliographystyle{IEEEtran}  
\bibliography{references}     

\end{document}